\documentclass[]{article}

\usepackage[round]{natbib}
\usepackage{authblk}

\usepackage[utf8]{inputenc} 
\usepackage[T1]{fontenc}    
\usepackage{hyperref}       
\usepackage{url}            
\usepackage{booktabs}       
\usepackage{amsfonts}       
\usepackage{nicefrac}       
\usepackage{microtype}      

\usepackage{amsmath}
\usepackage{mathtools}
\usepackage{bbm}

\usepackage{color}

\usepackage{amsthm}

\usepackage[capitalise]{cleveref}
\usepackage{subcaption}
\usepackage{array}

\definecolor{orange}{RGB}{255,153,0}
\definecolor{green}{RGB}{9,112,84}
\definecolor{yellow}{RGB}{52,100,100}
\definecolor{blue}{RGB}{101,153,255}

\newcommand{\indep}{\!\perp \!\!\! \perp\!}

\usepackage{times}
\usepackage{graphicx} 

\usepackage{tikz}
\usetikzlibrary{tikzmark}

\usepackage[ruled,vlined,linesnumbered]{algorithm2e}

\title{On the Difference Between the Information Bottleneck and the Deep Information Bottleneck}
\author{Aleksander Wieczorek}
\author{Volker Roth}
\affil{Department of Mathematics and Computer Science, University of Basel, Switzerland\\
	\texttt{\{aleksander.wieczorek, volker.roth\}@unibas.ch}}
\date{\today}

\begin{document}
	\maketitle
	\begin{abstract} 
		Combining the Information Bottleneck model with deep learning by replacing mutual information terms with deep neural nets has proved successful in areas ranging from generative modelling to interpreting deep neural networks. In this paper, we revisit the Deep Variational Information Bottleneck and the assumptions needed for its derivation. The two assumed properties of the data $X$, $Y$ and their latent representation $T$ take the form of two Markov chains $T-X-Y$ and $X-T-Y$. Requiring both to hold during the optimisation process can be limiting for the set of potential joint distributions $P(X,Y,T)$. We therefore show how to circumvent this limitation by optimising a lower bound for $I(T;Y)$ for which only the latter Markov chain has to be satisfied. The actual mutual information consists of the lower bound which is optimised in DVIB and cognate models in practice and of two terms measuring how much the former requirement $T-X-Y$ is violated. Finally, we propose to interpret the family of information bottleneck models as directed graphical models and show that in this framework the original and deep information bottlenecks are special cases of a fundamental IB model.
	\end{abstract} 
	\section{Introduction}
\label{sce:introduction}

Deep latent variable models such as generative adversarial networks~\citep{art:jsd:gan} or the variational autoencoder (VAE)~\citep{art:vae} have attracted much interest in the last few years. They have been used across many application areas and formed a conceptual basis for a number of extensions. One of popular deep latent variable models is the deep variational information bottleneck (DVIB)~\citep{art:deepIB:alemi}. Its foundational idea is that of applying deep neural networks to the information bottleneck (IB) model~\citep{art:tishby:ib} which finds a sufficient statistic $T$ of a given variable $X$ while retaining side information about a variable $Y$.

The original IB model as well as DVIB assume the Markov chain $T-X-Y$. Additionally, in the latter model, the Markov chain $X-T-Y$ appears by construction. The relationship between the two assumptions and how it influences the set of potential solutions have been neglected so far. In this paper, we clarify this relationship by showing that it is possible to lift the original IB assumption in the context of the deep variational information bottleneck. It can be achieved by optimising a lower bound on the mutual information between $T$ and $Y$ which follows naturally from the model's construction. This explains why DVIB can optimise over a set of distributions which is not overly restrictive.

This paper is structured as follows. In~\cref{sec:relatedWork} we describe the information bottleneck and deep variational information bottleneck models along with their extensions. \Cref{sec:discrepancy} introduces the lower bound on the mutual information which makes it possible to lift the original IB $T-X-Y$ assumption as well as the interpretation of this bound. It also contains the specification of IB as directed graphical model.  We provide concluding remarks in~\cref{sec:conclusion}.
	\section{Related Work on the Deep Information Bottleneck Model}
\label{sec:relatedWork}
The information bottleneck was originally introduced by~\citet{art:tishby:ib} as a compression technique in which a random variable $X$ is compressed while preserving relevant information about another random variable $Y$. The problem was originally formulated using only information theoretic concepts.  No analytical solution exists for the original formulation, however, an additional assumption that $X$ and $Y$ are jointly Gaussian distributed leads to a special case of the IB, the Gaussian information bottleneck, introduced by~\citet{art:chechik:gib}, where the optimal compression is also Gaussian distributed. The Gaussian information bottleneck has been further extended to sparse compression and to meta-Gaussian distributions (multivariate distributions with a Gaussian copula and arbitrary marginal densities) by~\citet{art:rey:sgib}. The idea of applying deep neural networks to model the information common to $X$ and $T$ as well as $Y$ and $T$ has resulted in the formulation of the deep variational information bottleneck~\citep{art:deepIB:alemi}. This model has been extended to account for invariance to monotonic transformations of the input variables by~\citet{art:deepCopulaIB}.

The information bottleneck method has also recently been applied to the analysis of deep neural networks~\citep{art:tishby:ib_deep}, by quantifying mutual information between the network layers and deriving an information theoretic limit on DNN efficiency. This has lead to attempts at explaining the behaviour of deep neural networks with the IB formalism~\citep{art:tishby:deep:theory:1,art:deep:theory:2}.

We now proceed to formally define the IB and DVIB models.

\paragraph*{Notation.} Throughout this paper, we adopt the following notation.
Define the \textit{Kullback-Leibler divergence} between two (discrete or continuous) probability distributions $P$ and $Q$ as $D_{KL} (P(X)\ ||\ Q(X)) = \mathbb{E}_{P(X)} \log \frac{P(X)}{Q(X)}$.
Note that the KL divergence is always non-negative.
The \textit{mutual information} between $X$ and $Y$ is defined as
\begin{equation}
I(X;Y) = D_{KL}\Big(P(X,Y)\ ||\ P(X)P(Y)\Big).
\label{eq:mi}
\end{equation}
Since the KL divergence is not symmetric, the divergence between the product of the marginals and the joint distribution has also been defined as the \textit{lautum information}~\citep{art:lautum}:
\begin{equation}
L(X;Y) = D_{KL}\Big(P(X)P(Y)\ ||\ P(X,Y)\Big).
\label{eq:lautum}
\end{equation}
Both quantities have conditional counterparts:
\begin{equation}
\begin{aligned}
	I(X;Y|Z) &= D_{KL}\Big(P(X,Y,Z)\ ||\ P(X|Z)P(Y|Z)P(Z)\Big)\\
	 &= \mathbb{E}_{P(Z)}D_{KL}\Big(P(X,Y|Z)\ ||\ P(X|Z)P(Y|Z)\Big),\\
	L(X;Y|Z) &= D_{KL}\Big(P(X|Z)P(Y|Z)P(Z)\ ||\ P(X,Y,Z)\Big)\\
	 &= \mathbb{E}_{P(Z)}D_{KL}\Big(P(X|Z)P(Y|Z)\ ||\ P(X,Y|Z)\Big).
\end{aligned}
\end{equation}

Let $H\left[ X\right] = -\mathbb{E}_{P(X)}\left[\log P(X)\right]$ denote \textit{entropy} for discrete and \textit{differential entropy} for continuous $X$.
Analogously, $H\left[ P(X|Y) \right] = -\mathbb{E}_{P(X,Y)}\left[\log P(X|Y)\right]$ denotes \textit{conditional entropy} for discrete and \textit{conditional differential entropy} for continuous $X$ and $Y$.

\subsection{Information Bottleneck}
\label{sec:relatedWork:sub:ib}

Given two random vectors $X$ and $Y$, the {Information Bottleneck} method~\citep{art:tishby:ib} searches for a third random vector $T$ which, while compressing $X$, preserves information contained in $Y$. The resulting variational problem is defined as follows:
\begin{equation}
\mbox{min}_{P(T|X)} I(X;T) - \beta I(T;Y)
\label{eq:ib}
\end{equation}
where $\beta$ is a parameter defining the trade-off between compression of $X$ and preservation of $Y$. 
The solution is the optimal conditional distribution of $T|X$. No analytical solution exists for the general IB problem defined by \cref{eq:ib}, however for discrete $X$ and $Y$ a numerical approximation of the optimal distribution $T$ can be found with the Blahut-Arimoto algorithm for rate-distortion function calculation~\citep{art:tishby:ib}. Note that the assumed property $T-X-Y$ of the solution is used in the derivation of the model.

\paragraph*{Gaussian Information Bottleneck.}
For Gaussian distributed $(X,Y)$, let the partitioning of the joint covariance matrix be denoted as follows:
\begin{equation}
(X, Y) \sim \mathcal{N}
\left( 0, 
\begin{pmatrix*}[l]
\Sigma_{X} & \Sigma_{XY}\\
{\Sigma_{XY}}' & \Sigma_{Y}\\
\end{pmatrix*}
\right)
\label{eq:gib:distr}
\end{equation}
The assumption that $X$ and $Y$ are jointly Gaussian distributed leads to the Gaussian information bottleneck~\citep{art:chechik:gib} where the solution $T$ of~\cref{eq:ib} is also Gaussian distributed. 
$T$ is then a noisy linear projection of $X$, i.e.\ $T=AX+\xi$, where $\xi \sim \mathcal{N}(0, \Sigma_{\xi})$ is independent of $X$. This means that $T\sim \mathcal{N}(0,A\Sigma_{X} A^{T} + \Sigma_{\xi})$. The IB optimisation problem defined in \cref{eq:ib} becomes an optimisation problem over the matrix $A$ and noise covariance matrix $\Sigma_{\xi}$:
\begin{equation}
\mbox{min}_{A, \Sigma_{\xi}} I(X;AX+\xi) - \beta I(AX+\xi;Y)
\label{eq:gib}
\end{equation}
Recall that for $n$-dimensional Gaussian distributed random variables, entropy, and hence mutual information, have the following form:
$I(X; Y) = H(X) - H(X|Y) = \frac{1}{2}\log \left( (2\pi e)^n |\Sigma_{X}| \right) - \frac{1}{2}\log \left( (2\pi e)^n |\Sigma_{X|Y}| \right),$
where $\Sigma_{X}$ and $\Sigma_{X|Y}$ denote covariance matrices of $X$ and $X|Y$, respectively.
The Gaussian information bottleneck problem has an analytical solution, given by~\citet{art:chechik:gib}: for a fixed $\beta$,~\cref{eq:gib} is optimised by $\Sigma_{\xi} = I$ and $A$ having an analytical form depending on $\Sigma_{X}$ and eigenvectors and eigenvalues of $\Sigma_{X|Y}\Sigma_{Y}^{-1}$. Here again, the $T-X-Y$ assumption is used in the derivation of the solution.

\paragraph*{Sparse Gaussian Information Bottleneck.}
Sparsity of the compression in the Gaussian IB can be ensured by requiring the projection matrix $A$ to be diagonal, i.e.\ $A = \mbox{diag}(a_1, \dots , a_n)$. It has been shown by~\citep{art:rey:sgib} that, since  $\log |A\Sigma A^{T} +I| = \log |\Sigma A^{T}A +I|$ for any positive definite $\Sigma$ and symmetric $A$, the sparsity requirement simplifies \cref{eq:gib} to minimisation over diagonal matrices with positive entries $D=A^{T}A = \mbox{diag}({a_1}^{2}, \dots , {a_n}^{2}) = \mbox{diag}({d_1}, \dots , {d_n})$, i.e.:
\begin{equation}
\underset{D =\mbox{diag}({d_1}, \dots , {d_n})}{\mbox{min}} I(X;AX+\xi) - \beta I(AX+\xi;Y)
\label{eq:sgib}
\end{equation}
with $d_i = {a_i}^{2}$ and $\xi \sim \mathcal{N}(0, I)$ independent of $X$. 

\subsection{Deep Variational Information Bottleneck}
\label{sec:relatedWork:sub:dib}

The deep variational information bottleneck~\citep{art:deepIB:alemi} is a variational approach to the problem defined in \cref{eq:ib}. The main idea is to parametrise the conditionals  $ P(T|X) $ and $ P(Y|T)$ with neural networks so that the two mutual informations in \cref{eq:ib} can be directly recovered from two deep neural nets. To this end, one can express the mutual informations as follows:

\begin{equation}
\begin{aligned}
	I(X;T) &=\ D_{KL}\left( P(T|X) P(X) \| P(T) P(X)\right)\\
	&=\ \int P_\Phi(T|X) P(X) \log \frac{P(T|X)}{ P(T) } \, \mathrm{d}x\,\mathrm{d}t\\
	&=\ \mathbb{E}_{P(X)}  D_{KL}\left(P(T|X)\| P(T) \right)
\end{aligned}
\label{eq:ib:enc}
\end{equation}

\begin{equation}
\begin{aligned}
I(T;Y) &=\ D_{KL}\left(\left[\int P(T|Y,X)P(Y,X)\, \mathrm{d}x \right] \| P(T) P(Y)\right)\\
&=\ \int P(T|X,Y) P(X,Y) \log \frac{P(Y|T) P(T)}{ P(T) P(Y)} \, \mathrm{d}t\,\mathrm{d}x\,\mathrm{d}y\\
&=\ \mathbb{E}_{P(X,Y)}  \left[ \int  P(T|X,Y) \log P(Y|T) \,\mathrm{d}t \right] \\
&\quad - \mathbb{E}_{P(X,Y)} \left[ \log P(Y) \int P(T|X,Y)  \,\mathrm{d}t\right] \\
&=\ \mathbb{E}_{P(X,Y)} \mathbb{E}_{P(T|X,Y)}\log P(Y|T)  + H(Y), \\
&=\ \mathbb{E}_{P(X,Y)} \mathbb{E}_{P(T|X)}\log P(Y|T)  + H(Y),
\end{aligned}
\label{eq:ib:dec}
\end{equation}
where the last equality in \cref{eq:ib:dec} follows from the Markov assumption $T-X-Y$ in the information bottleneck model: $P(T|X,Y) =  P(T|X)$. The conditional $Y|T$ is computed by sampling from the latent representation $T$ as in the variational autoencoder~\citep{art:vae}.
Note that this form of the DVIB makes sure that one is only required to sample from the data distribution $P(X,Y)$, the variational decoder $P_\theta(Y|T)$, and the stochastic encoder $P_\phi(T|X)$, implemented as deep neural networks parametrised by $\theta$ and $\phi$, respectively. In the latter, $T$ depends only on $X$ because of the $T-X-Y$ assumption.

\paragraph*{Deep Copula Information Bottleneck.}
\citet{art:deepIB:alemi} argue that the entropy term $H(Y)$ in the last line of \cref{eq:ib:dec} can be omitted, as $Y$ is a constant. It has however been pointed out~\citep{art:deepCopulaIB} that the IB solution should be invariant to monotonic transformations of both $X$ and $Y$, since the problem is defined only in terms of mutual informations which exhibit such invariance (i.e.\ $I(X;T) = I(f(X);T)$ for an invertible $f$). The term remaining in \cref{eq:ib:dec} after leaving out $H(Y)$ does not have this property. Furthermore, problems limiting the DVIB when specifying marginal distributions of $T|X$ and $Y|T$ in \cref{eq:ib:enc,eq:ib:dec} have been identified~\citep{art:deepCopulaIB}. These considerations have lead to the formulation of the deep copula information bottleneck, where the data are subject to the following transformation $\tilde{X} = \Phi^{-1}(\hat{F}(X))$, where $\Phi$ and $\hat{F}$ are the Gaussian and empirical cdfs, respectively. This transformation makes them depend only on their copula and not on the marginals. This has also been shown to result in superior interpretability and disentanglement of the latent space $T$.

\subsection{Bounds on Mutual Information in Deep Latent Variable Models}
\label{sec:relatedWork:sub:vi}
The deep information bottleneck model can be thought of as an extension of the VAE. Indeed, one can incorporate a variational approximation $Q(Y|T)$ of the posterior $P(Y|T)$ to \cref{eq:ib:dec} and by $D_{KL}(Q(Y|T)\ ||\ P(Y|T)) \geq 0$ obtain $I(T;Y) \geq \mathbb{E}_{P(X,Y)} \mathbb{E}_{P(T|X)}\log Q(Y|T)  + H(Y)$~\citep{art:deepIB:alemi}.
A number of other bounds and approximations of mutual information have been considered in the literature. Many of them are motivated by obtaining a better representation of the latent space $T$. \citet{art:alemi:elbo} consider different encoding distributions $Q(T|X)$ and derive a common bound for $I(X;T)$ on the rate-distortion plane. They subsequently extend this bound to the case where it is independent of the sample, which makes it possible to compare VAE and GANs~\citep{art:alemi:gilbo}.

\citet{art:tishby:gaussianBound} use a Gaussian relaxation of the mutual information terms in the information bottleneck to bound them from below. They then proceed to compare the resulting method to Canonical Correlation Analysis~\citep{art:cca}.

Extensions of generative models with an explicit regularisation in the form of a mutual information term have been proposed~\citep{art:infovae,art:infogan}. In the latter, an explicit lower bound on the mutual information between the latent space $T$ and the generator network is derived.

Similarly, implicit regularisation of generative models in the form of dropout has been shown to be equivalent to the deep information bottleneck model~\citep{art:informationDropout:1,art:informationDropout:2}. The authors also mention that both Markov properties should hold in the IB solution as well as note that $T-X-Y$ is enforced by construction while $X-T-Y$ is only approximated by the optimal joint distribution of $X$, $Y$ and $T$. They do not, however, analyse the impact of both Markov assumptions and the relationship between them.


	\section{The Difference Between Information Bottleneck Models}
\label{sec:discrepancy}

In this section, we focus on the difference between the original and deep IB models. First, we examine how the different Markov assumptions lead to different forms that the $I(Y;T)$ term admits. In~\cref{sec:discrepancy:sub:2}, we consider both models and show that describing them as directed graphical models makes it possible to elucidate a fundamental property shared by all IB models. We then proceed to summarise the comparison in~\cref{sec:discrepancy:sub:3}.

\subsection{Clarifying the Discrepancy between the Assumptions in IB and DVIB}
\label{sec:discrepancy:sub:clarifying}

\begin{figure}[!ht]
	\centering
	\begin{subfigure}[b]{0.45\textwidth}
		\centering
		\includegraphics[width=0.5\textwidth]{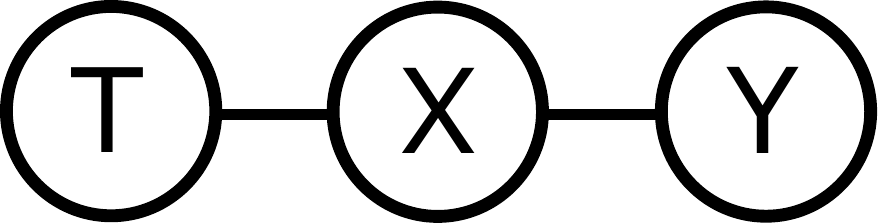}
		\caption{The original IB assumption.}
		\label{fig:ib:markov:1}
	\end{subfigure}
	\begin{subfigure}[b]{0.45\textwidth}
		\centering
		\includegraphics[width=0.5\textwidth]{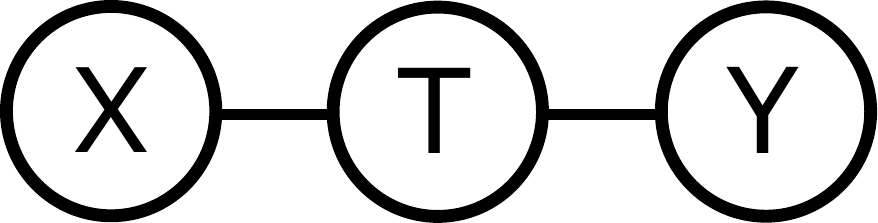}
		\caption{The DVIB assumption.}
		\label{fig:ib:markov:2}
	\end{subfigure}
	\caption{Markov assumptions for the Information Bottleneck and the Deep Information Bottleneck.}
	\label{fig:ib:markov}
\end{figure}

\paragraph*{Motivation.} The derivation of the deep variational information bottleneck model described in~\cref{sec:relatedWork:sub:dib} uses the Markov assumption $T-X-Y$ (last line of \cref{eq:ib:dec}, \cref{fig:ib:markov:1}). At the same time, by construction, the model adheres to the data generating process described by the following structural equations ($\eta_T, \eta_Y$ are noise terms independent of $X$ and $T$, respectively):
\begin{equation}
	\begin{aligned}
		T &= f_T(X, \eta_T),\\
		Y &= f_Y(T, \eta_Y).
	\end{aligned}
	\label{eq:ib:sem}
\end{equation}

This implies that the Markov chain $X-T-Y$ is satisfied in the model, too (\cref{fig:ib:markov:2}). Requiring that both Markov chains hold in the resulting joint distribution $P(X,Y,T)$ can be overly restrictive (note that no DAG with 3 vertices to which such distribution is faithful exists). Thus, the question arises if the $T-X-Y$ property in DVIB can be lifted. In what follows we show that it is indeed possible.

Recall from \cref{sec:relatedWork:sub:dib} that the DVIB model relies on sampling only from the data $P(X,Y)$, encoder $P(T|X)$ and decoder $P(Y|T)$. Therefore, for optimising the latent IB, we want to avoid specifying the full conditional $P(T|X,Y)$, since this would require us to explicitly model the joint influence of both $X$ and $Y$ on $T$ (which might be a complicated distribution).
We now proceed to show how to bound $I(T;Y)$ in a way that only involves sampling from the encoder $P(T|X)$ and circumvents modelling $P(T|X,Y)$ without using the $T-X-Y$ assumption.

\paragraph*{Bound derivation.} First, adopt the mutual information $I(T;Y)$ from the penultimate line of \cref{eq:ib:dec} (i.e.\ without assuming the $T-X-Y$ property):
\begin{equation}
I(T;Y) =\ \mathbb{E}_{P(X,Y)} \mathbb{E}_{P(T|X,Y)}\log P(Y|T)  + H(Y)
\label{eq:ib:dec:short}
\end{equation}
Now, rewrite \cref{eq:ib:dec:short} using $X-T-Y$ (i.e.\ $X \indep Y\ |\ T$):
\begin{equation}
\begin{aligned}
I(T;Y) &= \mathbb{E}_{P(X,Y)} \mathbb{E}_{P(T|X,Y)} \log P(Y|T)  + H(Y)\\ 
&=  \mathbb{E}_{P(X)} \mathbb{E}_{P(Y|X)} \mathbb{E}_{P(T|X,Y)} \log P(Y|T,X)  + H(Y).
\end{aligned}
\label{eq:der:1}
\end{equation}
Focusing on $\mathbb{E}_{P(Y|X)}\mathbb{E}_{P(T|X,Y)} \log P(Y|T,X)$ in \cref{eq:der:1}, the we obtain:
\begin{equation}
\begin{aligned}
\mathbb{E}&_{P(Y|X)}\mathbb{E}_{P(T|X,Y)} \log P(Y|T,X)= \int\!\int P(T,Y|X)\log P(T,Y|X) \mathrm{d}t\mathrm{d}y\\
&=\int\!\int P(T,Y|X) \log \frac{P(Y|X)P(T,Y|X)}{P(Y|X)P(T|X)} \mathrm{d}y\mathrm{d}t\\
&=D_{KL}\Big(P(Y,T|X) \| P(Y|X)P(T|X)\Big) + \int\!\int P(T,Y|X)\log P(Y|X) \mathrm{d}t\mathrm{d}y\\
&=D_{KL}\Big(P(Y,T|X) \| P(Y|X)P(T|X)\Big) + \int P(Y|X)\log P(Y|X) \mathrm{d}y\\
&=D_{KL}\Big(P(Y,T|X) \| P(Y|X)P(T|X)\Big) + \int\!\int P(Y|X)P(T|X)\log P(T|X) \mathrm{d}t\mathrm{d}y\\
&=D_{KL}\Big(P(Y,T|X) \| P(Y|X)P(T|X)\Big)\\
&\quad\quad\quad + \int\!\int P(Y|X)P(T|X)\log \frac{P(T|X)P(Y|X)P(T,Y|X)}{P(T,Y|X)P(T|X)} \mathrm{d}t\mathrm{d}y\\
&=D_{KL}\Big(P(Y,T|X) \| P(Y|X)P(T|X)\Big) + D_{KL}\Big( P(Y|X)P(T|X) \| P(Y,T|X)\Big)\\
&\quad\quad\quad + \mathbb{E}_{P(T|X)P(Y|X)}\log P(Y|T,X) \\
&\geq \mathbb{E}_{P(T|X)P(Y|X)}\log P(Y|T,X).
\end{aligned}
\label{eq:der:2}
\end{equation}	
Plugging \cref{eq:der:2} into \cref{eq:der:1}, i.e.\ averaging over $X$, and using $X \indep Y\ |\ T$ again, we arrive at:
\begin{equation}
\begin{aligned}
I(T;Y) &= \mathbb{E}_{P(X)} \mathbb{E}_{P(Y|X)} \mathbb{E}_{P(T|X,Y)}\log P(Y|T) + H(Y)\\
&= \mathbb{E}_{P(X)} \mathbb{E}_{P(Y|X)} \mathbb{E}_{P(T|X)} \log P(Y|T)\\
&\qquad+  \mathbb{E}_{P(X)}D_{KL} (P(Y,T|X) \| P(Y|X)P(T|X))\\
&\qquad+ \mathbb{E}_{P(X)}D_{KL} ( P(Y|X)P(T|X)\| P(Y,T|X) ) + H(Y)\\
&= \mathbb{E}_{P(X)} \mathbb{E}_{P(Y|X)} \mathbb{E}_{P(T|X)} \log P(Y|T)\\
&\qquad+ I(Y;T|X) + L(Y;T|X) + H(Y)\\
&\geq \mathbb{E}_{P(X)} \mathbb{E}_{P(Y|X)}  \mathbb{E}_{P(T|X)} \log P(Y|T)  +  H(Y).
\end{aligned}
\label{eq:der:3}
\end{equation}

\paragraph*{Interpretation.} According to \cref{eq:der:3}, the mutual information $I(T;Y)$ consists of 3 terms: its lower bound $\mathbb{E}_{P(X)} \mathbb{E}_{P(Y|X)}  \mathbb{E}_{P(T|X)} \log P(Y|T)$ which is actually optimised in DVIB and its extensions, $I(Y;T|X) + L(Y;T|X)$ which are $0$ when both Markov assumptions are satisfied, and the entropy term $H(Y)$.

\Cref{eq:der:3} shows how to bound the mutual information term $I(T;Y)$ in the IB model (\cref{eq:ib}) so that the value of the bound depends on the data $P(X,Y)$ and marginals $T|X$, $Y|T$ without using the Markov assumption $T-X-Y$. If we now again implement the marginal distributions as deep neural nets $P_\phi(T|X)$ and $P_\theta(Y|T)$, \cref{eq:der:3} provides the lower bound which is actually optimised in DVIB (\cref{eq:ib:dec}). During optimisation, the parameters $\phi$ in $P_\phi(T|X)$ and $\theta$ in $P_\theta(Y|T)$ are adjusted such that both $I(Y;T|X)$ and $L(Y;T|X)$ become small.
The terms $I(Y;T|X)$ and $L(Y;T|X)$ can thus be interpreted as a measure of how much the original IB assumption $T-X-Y$ is violated during the training of the model that implements $X-T-Y$ by construction.

The difference between the original IB and DVIB is that in the former, $T-X-Y$ is used to derive the general form of the solution $T$ while $X-T-Y$ is approximated as closely as possible by $T$ (as noted by~\citet{art:informationDropout:1}. In the latter, $X-T-Y$ is forced by construction, and $T-X-Y$ is approximated by optimising the lower bound given by~\cref{eq:der:3}. The 'distance' to a distribution satisfying both assumptions is measured by the tightness of the bound.

\subsection{The Original IB Assumption Revisited}
\label{sec:discrepancy:sub:2}


\paragraph*{Motivation for conditional independence assumptions in information bottleneck models.} In the original formulation of the information bottleneck (\cref{sec:relatedWork:sub:ib,eq:ib}), given by: $\mbox{min}_{P(T|X)} I(X;T) - \beta I(T;Y)$, one optimises over $P(T|X)$ while disregarding any dependence on $Y$. This suggests that the defining feature of the IB model is the absence of a direct functional dependence of $T$ on $Y$. This can be achieved e.g.\ by
the first structural equation in~\cref{eq:ib:sem}:
\begin{equation}
T = f_T(X, \eta_T).
\label{eq:sem:1}
\end{equation}
It means that any influence of $Y$ on $T$ must go through $X$. Note that this is implied by the original IB assumption $T-X-Y$, but not the other way around. In particular, the model given by $X-T-Y$ can also be parametrised such that there is no direct dependence of $T$ on $Y$, e.g.\ as in~\cref{eq:ib:sem}. This means that DVIB, despite optimising a lower bound on the IB, implements the defining feature of IB as well.

\begin{table}
	\caption{Directed graphical models of the information bottleneck.}
	\begin{small}
		\begin{tabular}{m{1.5cm}m{3.15cm}m{3.15cm}m{3.15cm}}
			\hline 
			Defining Markov assumption & \hspace{2mm}\includegraphics[width=0.21\textwidth]{figures/IB_Markov1.pdf} & \hspace{2mm}\includegraphics[width=0.21\textwidth]{figures/IB_Markov2.pdf} &  \hspace{7mm}Other / None \\ 
			\hline 
			Admissible DAG models &
			\begin{equation*}
			\begin{aligned}
			&T \rightarrow X \rightarrow  Y\\
			&T \leftarrow X \rightarrow  Y\\
			&T \leftarrow X \leftarrow  Y
			\end{aligned}
			\end{equation*}
			&
			\begin{equation*}
			\begin{aligned}
			&T\tikzmark{70} \leftarrow X \ \ \quad  \tikzmark{71}Y\\
			&T\tikzmark{60} \rightarrow X \ \ \quad  \tikzmark{61}Y
			\end{aligned}
			\end{equation*}
			&
			\begin{equation*}
			\begin{aligned}
			&T\tikzmark{00} \rightarrow X \rightarrow  \tikzmark{01}Y\\
			&T\tikzmark{10} \leftarrow X \rightarrow  \tikzmark{11}Y\\
			&T\tikzmark{20} \rightarrow X \leftarrow  \tikzmark{21}Y\\
			&T\tikzmark{30} \leftarrow X \leftarrow  \tikzmark{31}Y\\
			&T\tikzmark{50} \ \ \quad X \leftarrow  \tikzmark{51}Y
			\end{aligned}
			\end{equation*}
			\\ 
			\hline
		\end{tabular}
	\end{small}
	\label{table:ib_models}
	\begin{tikzpicture}[remember picture, overlay, bend left=15, -{>[scale=1.0]}]
	\draw ([yshift=2ex]pic cs:00) to ([yshift=2ex]pic cs:01);
	\draw ([yshift=2ex]pic cs:10) to ([yshift=2ex]pic cs:11);
	\draw ([yshift=2ex]pic cs:20) to ([yshift=2ex]pic cs:21);
	\draw ([yshift=2ex]pic cs:30) to ([yshift=2ex]pic cs:31);
	\draw ([yshift=2ex]pic cs:50) to ([yshift=2ex]pic cs:51);
	\draw ([yshift=2ex]pic cs:60) to ([yshift=2ex]pic cs:61);
	\draw ([yshift=2ex]pic cs:70) to ([yshift=2ex]pic cs:71);
	\end{tikzpicture}
\end{table}

\paragraph*{Information bottleneck as a directed graphical model.} The above discussion leads to the conclusion that the IB assumptions might also be described by directed graphical models. Such models encode conditional independence relations with d-separation (for the definition and examples of d-separation in DAGs, see~\citet{book:gm:lauritzen} or~\citet[Chapters~1.2.3~and~11.1.2]{book:pearl:causality}). In particular, any pair of variables d-separated by $Z$ is conditionally independent given $Z$. The arrows of the DAG are assumed to correspond to the data generating process described by a set of structural equation (as in \cref{eq:ib:sem}). Therefore, the following probability factorisation and data generating process hold for a DAG model:
\begin{equation}
P(X_1,X_2,\dots,X_n) = \prod_{i} P(X_i|pa(X_i))
\label{eq:dag:factorisation}
\end{equation}
\begin{equation}
X_i = f_i(pa(X_i), U_i),
\label{eq:dag:dgp}
\end{equation}
where $pa(X_i)$ stands for the set of direct parents of $X_i$ and $U_i$ are exogenous noise variables.

Let us now focus again on the motivation for the $T-X-Y$ assumption in \cref{eq:ib}. It prevents the model to choose a degenerate solution of $T=Y$ (in which case $I(X;T)=\mbox{const.}$ and $I(T;Y)=\infty$). Note, however, that while $T-X-Y$ is a sufficient condition for such solution to be excluded (which justifies the correctness of the original IB), the necessary condition is that $T$ cannot depend directly on $Y$. This means that the IB Markov assumption can be indeed reduced to requiring the absence of a direct arrow from $Y$ to $T$ in the underlying DAG. Note that this can be achieved in the undirected $X-T-Y$ model, too. One thus wishes to avoid degenerate solutions which impair the bottleneck nature of $T$: it should contain information about both $X$ and $Y$, the trade-off between them being steered by $\beta$. It is therefore necessary to exclude DAG structures which encode independence of $X$ and $T$ as well as $Y$ and $T$. Such independences are achieved by collider structures in DAGs, i.e.\ $T \rightarrow Y \leftarrow X$ and $T \rightarrow X \leftarrow Y$ (they lead to degenerate solutions of $I(X;T)=0$ and $I(T;Y)=0$, respectively). To sum up, the goal of asserting the conditional independence assumption in~\cref{eq:ib} is to avoid degenerate solutions which impair the bottleneck nature of the representation $T$. When modelling the information bottleneck with DAG structures, one has to exclude ate arrow $Y\rightarrow T$ as well as collider structures. A simple enumeration of the possible DAG models for the information bottleneck results in 10 distinct models listed in~\cref{table:ib_models}.

As can be seen, considering the information bottleneck as a directed graphical model (DAG) makes room for a family of models which fall into 3 broad categories: satisfying one of the two undirected Markov assumptions $T-X-Y$ or $X-T-Y$, as described in~\cref{sec:discrepancy:sub:clarifying} or neither of them (see~\cref{table:ib_models}). The difference between particular models lies in the necessity to specify different conditional distributions and parametrising them, which might lead to situations in which no joint distribution $P(X,Y,T)$ exists (which is likely to be the case in the third category). Focusing on the two first categories, we see that the former corresponds to the standard parametrisations of the information bottleneck and the Gaussian information bottleneck (see~\cref{sec:relatedWork:sub:ib}). In the latter, we see the deep information bottleneck (\cref{eq:ib:sem}) as the first DAG. Note also that the second DAG satisfying the $X-T-Y$ assumption in~\cref{table:ib_models} defines the probabilistic CCA model~\citep{art:probcca}. This is not surprising, since the solution of CCA and the Gaussian information bottleneck use eigenvectors of the same matrix~\citep{art:chechik:gib}.



\subsection{Comparing IB and DVIB Assumptions}
\label{sec:discrepancy:sub:3}

\begin{table}[!ht]
	\caption{Comparison of the Information Bottleneck and Deep Variational Information Bottleneck.}
	\begin{small}
		\begin{tabular}{m{2.5cm}m{4.2cm}m{4.2cm}}
			\hline 
			& Information Bottleneck (IB) & Deep Information Bottleneck (DVIB) \\ 
			\hline 
			\hline
			Assumed Markov chain & \hspace{7mm}\includegraphics[width=0.21\textwidth]{figures/IB_Markov1.pdf} & \hspace{7mm}\includegraphics[width=0.21\textwidth]{figures/IB_Markov2.pdf} \\ 
			\hline 
			Possible set of structural equations &
			\begin{equation*}
			\begin{aligned}
			T &= f_T(X, \eta_T),\\
			Y &= f_Y(X, \eta_Y)
			\end{aligned}
			\end{equation*}
			&
			\begin{equation*}
			\begin{aligned}
			T &= f_T(X, \eta_T),\\
			Y &= f_Y(T, \eta_Y)
			\end{aligned}
			\end{equation*}			\\ 
			Corresponding DAG & \hspace{12mm}\includegraphics[width=0.14\textwidth]{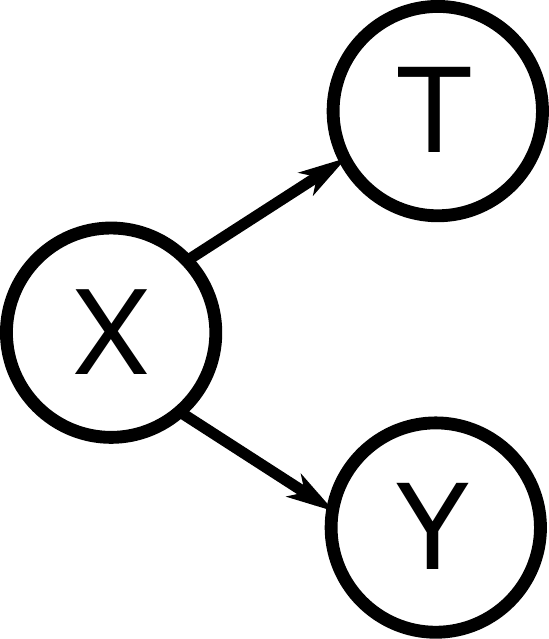} & \hspace{7mm}\includegraphics[width=0.21\textwidth]{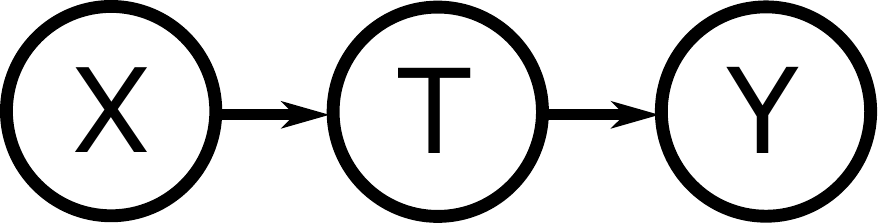} \\ 
			\hline
			Optimised term corresponding to $I(T;Y)$ &
			\begin{equation*}
			\begin{aligned}
			&\mathbb{E}_{P(X,Y)} \mathbb{E}_{P(T|X)} \log P(Y|T)\\ 
			&+ I(Y;T|X) + L(Y;T|X)\\
			&+ H(Y)
			\end{aligned}
			\end{equation*}
			&
			\begin{equation*}
			\begin{aligned}
			&\mathbb{E}_{P(X,Y)} \mathbb{E}_{P(T|X)} \log P(Y|T)\\
			&+ H(Y)
			\end{aligned}
			\end{equation*}
			\\ 
			\hline
		\end{tabular}
	\end{small}
	\label{table:in_comparison}
\end{table}

The original and deep information bottleneck models differ by using different Markov assumptions (see \cref{fig:ib:markov}) in the derivation of the respective solutions.
As demonstrated in \cref{sec:discrepancy:sub:clarifying}, DVIB optimises a lower bound on the objective function of IB. The tightness of the bound measures to what extent the IB assumption (\cref{fig:ib:markov:1}) is violated. As described in \cref{sec:discrepancy:sub:2}, characterising both models in as directed graphical models results in two different DAGs for the IB and DVIB. Both models are summarised in \cref{table:in_comparison}.

	\section{Conclusion}
\label{sec:conclusion}

In this paper, we show how to lift the information bottleneck Markov assumption $T-X-Y$ in the context of the deep information bottleneck model in which $X-T-Y$ holds by construction. 
This result explains why standard implementations of the deep information bottleneck can optimise over a larger amount of joint distributions $P(X,T,Y)$ while only specifying the marginal $T|X$. It is made possible by optimising the lower bound on the mutual information $I(T;Y)$ provided here rather than the full mutual information. We also provide a description of the information bottleneck as a DAG model and show that it is possible to identify a fundamental necessary feature of the IB in the language of directed graphical models. This property is satisfied in both the original and deep information bottleneck.

	\small{
		\bibliographystyle{plainnat}
		\bibliography{noteIB}
	}
\end{document}